\documentclass[runningheads]{llncs}
\usepackage{graphicx}

\usepackage{tikz}
\usepackage{comment}
\usepackage{amsmath,amssymb} 
\usepackage{color}
\usepackage{hyperref}

\begin{document}
\pagestyle{headings}
\mainmatter
\def\ECCVSubNumber{7426}  

\title{MarginDistillation: distillation for margin-based softmax} 

\author{Svitov David\inst{1,2} \and
Alyamkin Sergey\inst{1}}

\institute{Expasoft\\
\email{d.svitov@expasoft.tech} \and
Institute of Automation and Electrometry
of the Siberian Branch of the Russian Academy of Sciences}

\maketitle

\begin{abstract}
The usage of convolutional neural networks (CNNs) in conjunction with a margin-based softmax approach demonstrates a state-of-the-art performance for the face recognition problem. Recently, lightweight neural network models trained with the margin-based softmax have been introduced for the face identification task for edge devices. In this paper, we propose a novel distillation method for lightweight neural network architectures that outperforms other known methods for the face recognition task on LFW, AgeDB-30 and Megaface datasets. The idea of the proposed method is to use class centers from the teacher network for the student network. Then the student network is trained to get the same angles between the class centers and the face embeddings, predicted by the teacher network.
Implementation of our methods is available at: \\ \href{https://github.com/david-svitov/margindistillation}{https://github.com/david-svitov/margindistillation}
\keywords{distillation, margin-based softmax, ArcFace}
\end{abstract}

\section{Introduction}
The development of edge devices has sparked significant interest in lightweight face recognition access systems. This type of solution is based on optimized neural network architectures for mobile devices. A typical example of such network is MobileFaceNet \cite{chen2018mobilefacenets}, designed specifically for the face recognition on devices with low computing power. The usage of margin-base softmax approach \cite{deng2019arcface,liu2017sphereface,wang2018cosface} in the training procedure helps to obtain the state-of-the-art performance for the face recognition tasks. \\
Despite fast and compact mobile network architectures give lower face recognition accuracy than the full-size ones, in some applications, such as biometric access systems, it nevertheless plays a critical role. Distillation is a method that helps to achieve the highest accuracy for mobile neural network architectures, where  the knowledge is transferred from a heavy teacher network to a small student network. In this article we propose a novel distillation method called \textit{MarginDistillation} to reduce the gap between teacher and student networks during distillation process. \\
The idea of the proposed method is to copy class centers from a teacher network to a student network and freeze class centers for the whole distillation procedure, where the student network is trained to get angles between given class centers and face embeddings the same as in the teacher network. It allows the student network to better reproduce results of the teacher network trained with the margin-based loss function.\\
The main contributions of our work:
\begin{itemize}
  \item We have proposed a novel method for the distillation of neural networks trained with a margin-based softmax.
  \item The proposed method allows a gap reduction between teacher and student networks for face recognition problem. The accuracy of the mobile face recognition neural network achieved with our method exceeds other known distillation methods on different datasets: LFW \cite{huang2008labeled}, AgeDB-30 \cite{moschoglou2017agedb} and MageFace \cite{kemelmacher2016megaface} dataset.
  \item In presented work we made direct comparison of different distillation methods. Code for implemented methods and comparison experiments is available on the github.
\end{itemize}

\section{Related Works}
\textbf{Margin-based softmax.} 
There are several variations of the margin-based softmax used for the training of neural networks for the face recognition problem. They include Cosface \cite{wang2018cosface}, Sphereface \cite{liu2017sphereface}, and ArcFace \cite{deng2019arcface} approaches, which all can be described by the general formula:
\begin{equation}\label{margin_based} 
L = -\frac{1}{N} \sum_{i=1}^N log\frac{e^{s(cos(\theta_{y_i}m_1 + m_2) - m_3)}}{e^{s(cos(\theta_{y_i}m_1 + m_2) - m_3))} + \sum_{j=1, j \neq y_i}^n e^{s\:cos \theta_j}}.
\end{equation}
The listed methods are obtained from the formula \ref{margin_based} by the substitution of parameters. Sphereface: \begin{math}m_1=4, m_2=m_3=0\end{math}; Cosface: \begin{math}m_1=1, m_2=0, m_3=0.35\end{math}; Arcface: \begin{math}m_1=1, m_2=0.5, m_3=0\end{math}. The ArcFace approach for the face recognition task demonstrates the state-of-the-art performance on LFW, AgeDB-30 and MegaFace datasets.\\
\textbf{Distillation.} The knowledge distillation from the teacher network to the student network was proposed by Hinton \textit{et al.} \cite{hinton2015distilling}. It is an approach for training a small student neural network by transferring knowledge from a heavy teacher network. The key idea of the distillation proposed by Hinton is to transfer the knowledge about smoothed probability distribution of the output layer of the teacher network to the student network.\\
Some researchers continue to develop the idea of using a smoothed probability distribution as labels for training student network. For example, Fukuda \textit{et al.} \cite{fukuda2017efficient} proposed an approach to distill an ensemble of neural networks into a single student network. Sau \& Balasubramanian \cite{sau2016deep} proposed a regularization method that allows training a student network with a noisy teacher. Furlanello \textit{et al.}  \cite{furlanello2018born} trained the student network parameterized identically to the teacher network.\\
Another approach to the knowledge transfer is a distillation of hidden layers. Huang \textit{et al.} \cite{huang2017like} trained the student network to reproduce the distribution of weights on the hidden layers of the teacher network. Romero \textit{et al.} \cite{romero2014fitnets} used the outputs of intermediate layers of a teacher and a student network in a distillation procedure to regularize training. Chen \textit{et al.} \cite{chen2018learning} used a preservation of local object relationships for regularization. In their work \begin{math}L_2\end{math} distances between feature vectors of the student network are minimized, depending on the distance between corresponding vectors in the teacher network. Wonpyo \textit{et al.} \cite{park2019relational} proposed relational knowledge distillation that penalize structural differences in the samples relations.\\
For training lightweight face recognition neural networks with margin-based softmax, following distillation methods are used: Triplet distillation\cite{feng2019triplet}, Angular distillation \cite{duong2019shrinkteanet} and Margin Based Knowledge Distillation\cite{nekhaev2020margin}. \\
In \textit{Triplet distillation} approach, the student neural network is trained with a triplet loss function and margin, calculated based on the distances between the anchor and negative, the anchor and positive examples, predicted by the teacher network. \\
\textit{Angular distillation} approach minimizes the angle between teacher and student embedding vectors for each sample.\\
In \textit{Margin Based Knowledge Distillation} it is proposed to distill knowledge via smooth probability distribution obtained from the formula \ref{margin_based}, via dividing by the temperature value \begin{math}T\end{math}.\\

\section{Proposed Approach}
\subsection{Teacher and student networks}
The ResNet100 \cite{he2016deep} architecture was chosen as a teacher network. It has a large number of parameters and helps to achieve a high accuracy on face recognition tasks. The novel lightweight architecture called MobileFaceNet(ReLU) \cite{chen2018mobilefacenets} was used as the student network. In our experiments we made one modification of MobileFaceNet architecture: the dimension of the embedding vector was increased to 512 to make it compatible with ResNet-100 embedding. Table \ref{table_properties} shows comparison of parameters for the teacher and student networks.

\setlength{\tabcolsep}{4pt}
\begin{table}[h]
\begin{center}
\caption{Parameters of the considered networks. Network run time was measured for $112 \times 112 \times 3$ input images on a machine with a processor: Intel Xeon(R) CPU E3-1270 v3 @ 3.50GHz $\times$ 8.}
\begin{tabular}{lll}
\hline
                                                   & ResNet100 & MobileFaceNet \\ \hline
\multicolumn{1}{l|}{FLOPs / \begin{math}10^9\end{math}} & 24.2      & 0.44          \\
\multicolumn{1}{l|}{Size / Mb}                        & 261.2         & 5.3            \\
\multicolumn{1}{l|}{Number of parameters / \begin{math}10^6\end{math}}                        & 52.56 & 1.19              \\
\multicolumn{1}{l|}{Time / ms}                         & \begin{math} 401 \pm 25.7 \end{math} & \begin{math} 42.2 \pm 5.48 \end{math}\\ \hline            
\end{tabular}
\label{table_properties}
\end{center}
\end{table}
\setlength{\tabcolsep}{1.4pt}

\subsection{MarginDistillation}
Let \begin{math}x_{s_i}\in{R^D}\end{math} denotes the feature vector of the student network for the sample with number \begin{math}i\end{math}, \begin{math}x_{t_i}\in{R^D}\end{math} denotes the feature vector of the teacher network for the same sample. We will denote the weight matrices of the last layer of the student and teacher networks, respectively, by \begin{math}W_{s}\in{R^{D\times{n}}}\end{math} and \begin{math}W_{t}\in{R^{D\times{n}}}\end{math}. The column with index \begin{math}j\end{math} corresponding to the center of the class \begin{math}y_i\end{math} will be denoted by \begin{math}W_{s_j}\in{R^{D}}\end{math} and \begin{math}W_{t_j}\in{R^{D}}\end{math} for the student and teacher networks. \\
Methods based on adding the margin \begin{math}m\end{math} to the \begin{math}softmax\end{math} function, normalize the weight matrix and sample vectors by 1: \begin{math}||W_j|| = 1\end{math} and \begin{math}||x_i|| = 1\end{math}. This normalization allows to consider the output of the \begin{math}logit\end{math} layer as the cosine of the angles between sample vectors and corresponding class centers: \begin{math}{W_j}^T x_i = ||W_j||\cdot||x_i|| cos(\theta_j)=cos(\theta_j)\end{math}. We will consider ArcFace \cite{deng2019arcface} as a special case of the margin-based softmax approach, since it gives the best performance among the margin-based methods:
\begin{equation}\label{eq_arcface} 
L_{\textrm{ArcFace}} = -\frac{1}{N} \sum_{i=1}^N log\frac{e^{s(cos(\theta_{y_i} + m))}}{e^{s(cos(\theta_{y_i} + m))} + \sum_{j=1, j \neq y_i}^n e^{s\:cos \theta_j}}.
\end{equation} 
In ArcFace, margin \begin{math}m\end{math} is fixed at 0.5. We propose to distill  the knowledge from the teacher network by calculating the margin values \begin{math}m\end{math} for each sample \begin{math}i\end{math}. The proposed distillation method contains two key ideas:
\begin{itemize}
\item Class centers found by the teacher network are used for the student network: \begin{math}W_{s} = W_{t}\end{math}. Since class centers are learning values, a deeper network is able to learn more optimal position of classes on the hypersphere.
\item 
Calculated margin values \begin{math}m_i\end{math} are used for the distillation. They explicitly control the distance between vectors \begin{math}x_{s_i}\end{math} and corresponding class centers \begin{math}W_{s_j}\end{math}: larger \begin{math}m_i\end{math} leads to the stronger attraction of the vector \begin{math}x_{s_i}\end{math} to the class center.  It is proposed to calculate \begin{math}m_i\end{math} based on the information from the teacher.
\end{itemize}
\begin{figure}[h]
\center{\includegraphics[scale=0.45]{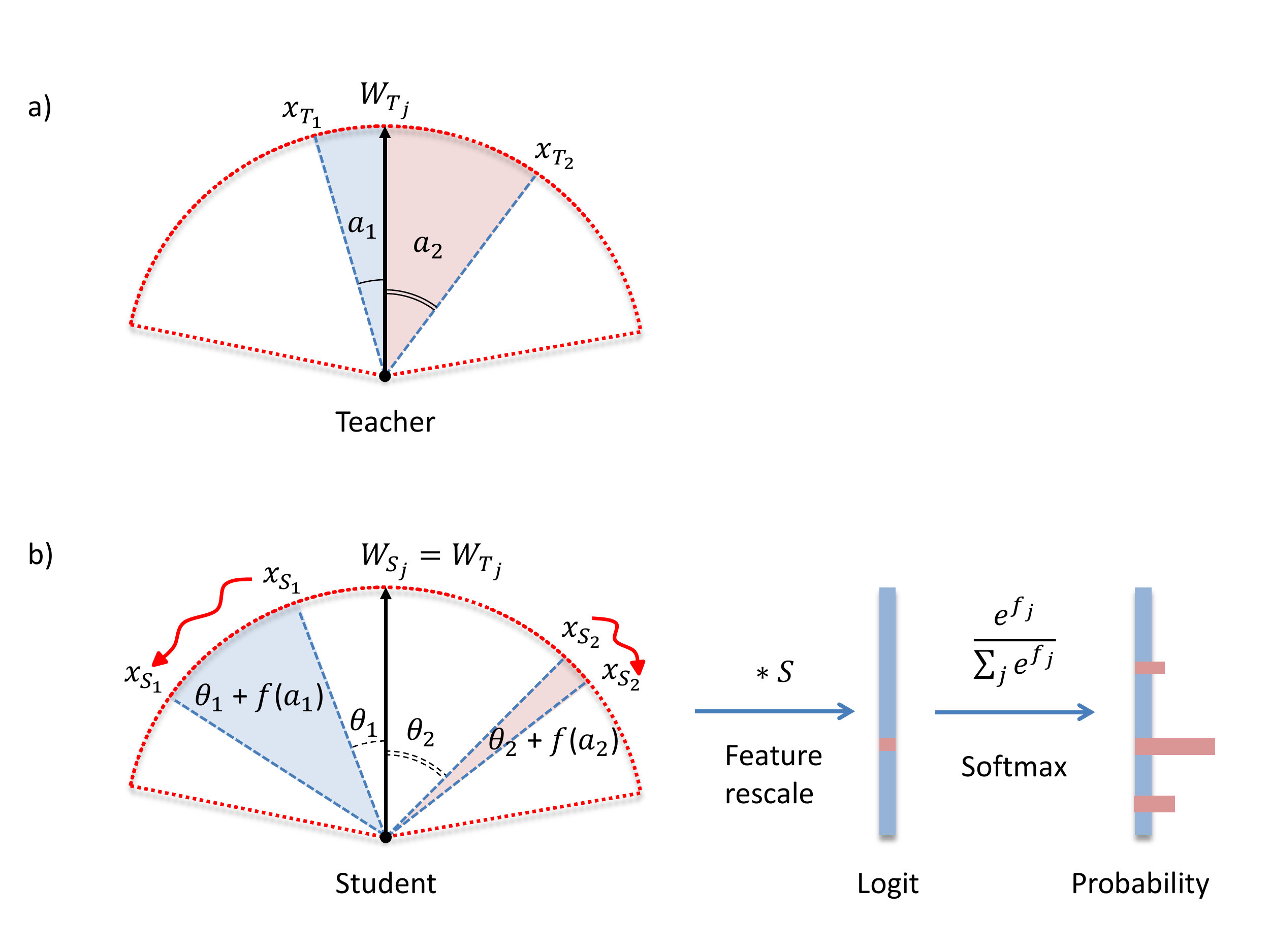}}
\caption{a) Using formula \ref{calc_angle} from the center of class $W_{t_j}$ and the sample vector $x_{t_i}$ of the teacher network, an angle $a_i$ is calculated. b) The angle $a_i$ is obtained from the teacher network and is used to calculate the margin $f(a_i)$ according to the formula \ref{calc_margin}. The smaller the angle between the sample vector and the center of the corresponding class in the teacher network, the greater the margin when training the student network. The margin calculated this way is used as $m$ in ArcFace (formula \ref{eq_arcface}). The larger the margin, the stronger the student network pulls the feature vector toward the center of the class.}
\label{fig:image}
\end{figure}
The intuition behind the proposed method is to pull feature vectors and class center closer to each other for the student network, when these vectors are close for the teacher network. It allows the knowledge transfer from the teacher to the student to be more efficient, because the student network focuses on samples with more confident predictions, while paying less attention to samples with low confidence.

Margins \begin{math}m_i\end{math} are calculated based on the angle between \begin{math}x_{t_i}\in{R^D}\end{math} and \begin{math}W_{t_j}\in{R^{D}}\end{math}. In other words, the margin values are calculated based on the angle between the center of the class \begin{math}y_i\end{math} and the sample vector \begin{math}i\end{math} in the teacher network.  Margin \begin{math}m\end{math} for the sample with index \begin{math}i\end{math} is calculated similarly to the triplet distillation margin as: 
\begin{equation}\label{calc_margin} 
m_i = \frac {m_{max} - m_{min}}{a_{max}} a_i + m_{min},
\end{equation} 
\begin{equation}\label{calc_angle} 
a_i = \frac {W_{t_j}^T x_{t_i}}{||W_{t_j}||\cdot||x_{t_i}||},
\end{equation} 
where we fix \begin{math}m_{max}=0.5\end{math} and \begin{math}m_{min}=0.2\end{math} - maximum and minimum margin values. And \begin{math}a_{max}\end{math} takes value of the largest angle \begin{math}a\end{math} in the mini-batch. \\
Our approach to the distillation allows transmitting the information about relative vectors position on the hypersphere without imposing strict limitation on the student feature vectors.

\section{Experiments}
\subsection{Implementation Details} 
\textbf{Pre-processing.} 
We use MTCNN method \cite{zhang2016joint} for the detection and alignment of a face in an image. For training procedure MS1MV2 \cite{deng2019arcface} was used. It is semi-automatic cleaned MS-Celeb-1M \cite{guo2016ms} dataset, proposed by ArcFace authors.\\
After the face alignment using the key points obtained by MTCNN, the images were cropped to \begin{math}112 \times 112\end{math} pixels. The pixel values were normalized to the range [-1, 1].\\
\textbf{Training.} 
In order to get a teacher network the ResNet100 was trained with ArcFace loss function. All distillation methods were compared in the same scenario, where the knowldege was transferred from the trained ResNet100 to MobileFaceNet(ReLU). We use the following setup for the distillation by our approach: mini-batch size is 512, learning rate was set to 0.1 and decreased 10 times by 100'000, 160'000 and 220'000 iterations. Optimization was performed by the SGD algorithm with a momentum of 0.9 and a weight decay of \begin{math}5e-4\end{math}. The values of the maximum and the minimum possible margins were fixed as: \begin{math}m_{max}=0.5\end{math} and \begin{math}m_{min}=0.2\end{math}. Scale factor \begin{math}s\end{math} was set to 64 as in ArcFace. For the training MarginDistillation, a modification of the official implementation of ArcFace on MXNet was used.\\
In order to compare MarginDistillation with other distillation methods for margin-based softmax proposed previously, we imlemented Triplet distillation \cite{feng2019triplet}, Angular distillation for feature direction \cite{duong2019shrinkteanet} and Margin Based Knowledge Distillation \cite{nekhaev2020margin} on MXNet. The methods were trained with parameters recommended in the corresponding papers. The source code can be found in the article repository.\\
\textbf{Evaluation.}
\textit{LFW, AgeDB-30:} The considered datasets are widely used in the tasks of evaluating facial verification algorithms. They contain about 3000 positive and 3000 negative pairs of examples. At the testing stage, the trained network was used to obtain a feature vector for a pre-processed image of the face and its horizontal flip copy. Both vectors were then concatenated. The resulting vector was used for the verification. The accuracy is measured as the percentage of correctly verified pairs of examples.\\
\textit{MegaFace:} It is the most representative and challenging open testing protocol for the face recognition task. MegaFace includes 1 million facial images for 690'000 people, as a sample for the formation of distractors, and 100'000 for 530 people from the FaceScrub \cite{ng2014data} dataset for identification. The measured metric is the top-1 accuracy for identification with 1 million distractors.\\
\subsection{Experimental results}
\setlength{\tabcolsep}{4pt}
\begin{table}[h]
\begin{center}
\caption{Verification accuracy at LFW and AgeDB-30. The experiments used the version of MobileFaceNet with ReLU.}
\begin{tabular}{lllll}
\hline
Architecture  & Training method          & LFW \% & AgeDB-30 \\ \hline
ResNet100     & ArcFace\cite{deng2019arcface}      & 99.76 & 98.21 \\
(teacher)     &                          &        &      \\
MobileFaceNet & ArcFace\cite{deng2019arcface}     & 99.51 & 96.13 \\ 
(student)     &                          &        &       \\ \hline
MobileFaceNet & triplet distillation L2\cite{feng2019triplet} & 99.56 & 96.23  \\ 
MobileFaceNet & triplet distillation cos\cite{feng2019triplet} & 99.55 & 95.60 \\ 
MobileFaceNet & margin based with T=4\cite{nekhaev2020margin}  & 99.41 & 96.01 \\ 
MobileFaceNet & angular distillation\cite{duong2019shrinkteanet} & 99.55 & 96.01 \\ \hline
MobileFaceNet & MarginDistillation (our)    & \textbf{99.61} & \textbf{96.55}    \\
\end{tabular}
\label{table_LFW}
\end{center}
\end{table}
\setlength{\tabcolsep}{1.4pt}
 
As shown in Table \ref{table_LFW}, the trained teacher network reaches 99.76\% on LFW and 98.21\% on AgeDB-30. The student network trained with ArcFace reaches 99.51\% on LFW and 96.13\% on AgeDB-30. Our approach gives the best results on AgeDB-30 reaching 96.55\% and on LFW reaching 99.61\%.

\setlength{\tabcolsep}{4pt}
\begin{table}[h]
\begin{center}
\caption{Identification accuracy using MegaFace protocol. With 1 million distractors. The experiments used the version of MobileFaceNet with ReLU.}
\begin{tabular}{lll}
\hline
Architecture   & Training method                 & MegaFace acc. \% \\ \hline
ResNet100      & ArcFace\cite{deng2019arcface}             & 98.35          \\
(teacher)      &                                 &                  \\
MobileFaceNet  & ArcFace\cite{deng2019arcface}             & 90.62          \\ 
(student)      &                                 &                   \\ \hline
MobileFaceNet  & triplet distillation L2\cite{feng2019triplet}  & 89.10     \\ 
MobileFaceNet  & triplet distillation cos\cite{feng2019triplet} & 86.52     \\
MobileFaceNet  & margin based with T=4\cite{nekhaev2020margin}  &  90.77     \\
MobileFaceNet  & angular distillation\cite{duong2019shrinkteanet}    &  90.73   \\ \hline
MobileFaceNet  & MarginDistillation (our)           & \textbf{91.70} \\
\end{tabular}
\label{table_megaface}
\end{center}
\end{table}
\setlength{\tabcolsep}{1.4pt}

Since many algorithms show high accuracy on LFW dataset, it cannot be used to conclude which algorithm is suitable for usage in real world scenarios. MegaFace dataset is more challenging, which includes a much larger number of people and images. On the MegaFace dataset the teacher network reaches 98.35\%. The student network trained with ArcFace reaches 90.62\%. As shown in Table \ref{table_megaface}  our method demonstrates the best accuracy of 91.70\%. In Triplet distillation methods, some drawdown of accuracy were noted, although they demonstrated good results on LFW.

\section{Conclusion}
A novel network distillation approach for the face recognition was introduced: MarginDistillation. We demonstrated the effectiveness of using the centers of classes from the teacher network and teacher dependent margin to distill networks with margin-based softmax. The proposed method was compared with other distillation methods and demonstrated superior performance on LFW, AgeDB-30 and challenging Megaface datasets.

\clearpage
%
%
\bibliographystyle{splncs04}
\bibliography{egbib}
\end{document}